\documentclass{llncs}

\usepackage[misc]{ifsym}
\usepackage[colorinlistoftodos]{todonotes}

\usepackage[caption=false]{subfig}
\usepackage{float}

\usepackage{multirow}
\usepackage{rotating}

\makeatletter
\def\Hline{%
\noalign{\ifnum0=`}\fi\hrule \@height 1pt \futurelet
\reserved@a\@xhline}
\makeatother

\begin{document}

\title{Infinite Brain MR Images: PGGAN-based Data Augmentation for Tumor Detection}
\titlerunning{GAN-based Brain MRI Data Augmentation}  
%
\author{Changhee Han\inst{1} (\Letter) \and Leonardo Rundo\inst{2, 3}  \and Ryosuke Araki\inst{4} \and Yujiro Furukawa\inst{5} \and Giancarlo Mauri\inst{2} \and Hideki Nakayama\inst{1} \and Hideaki Hayashi\inst{6}}
\authorrunning{Changhee Han et al.} 
%
%
\institute{Graduate School of Information Science and Technology,\\The University of Tokyo, Tokyo, Japan \\ 
\email{han@nlab.ci.i.u-tokyo.ac.jp} \\
\and Department of Informatics, Systems and Communication,\\University of Milano-Bicocca, Milan, Italy\\
\and
Institute of Molecular Bioimaging and Physiology (IBFM),\\Italian National Research Council (CNR), Cefal\`{u} (PA), Italy
\and
Graduate School of Engineering, Chubu University, Aichi, Japan
\and Kanto Rosai Hospital, Kanagawa, Japan
\and Department of Advanced Information Technology,\\Kyushu University, Fukuoka, Japan}
\maketitle              
\begin{abstract}
Due to the lack of available annotated medical images, accurate computer-assisted diagnosis requires intensive Data Augmentation (DA) techniques, such as geometric/intensity transformations of original images; however, those transformed images intrinsically have a similar distribution to the original ones, leading to limited performance improvement. To fill the data lack in the real image distribution, we synthesize brain contrast-enhanced Magnetic Resonance (MR) images---realistic but completely different from the original ones---using Generative Adversarial Networks (GANs). This study exploits Progressive Growing of GANs (PGGANs), a multi-stage generative training method, to generate original-sized $256\times256$ MR images for Convolutional Neural Network-based brain tumor detection, which is challenging \textit{via} conventional GANs; difficulties arise due to unstable GAN training with high resolution and a variety of tumors in size, location, shape, and contrast. Our preliminary results show that this novel PGGAN-based DA method can achieve promising performance improvement, when combined with classical DA, in tumor detection and also in other medical imaging tasks.

\end{abstract}

\section{Introduction}
Along with classical methods~\cite{Rundo,rundo2016WIRN}, Convolutional Neural Networks (CNNs) have dramatically improved medical image analysis~\cite{Bevilacqua,Brunetti}, such as brain Magnetic Resonance Imaging (MRI) segmentation~\cite{Havaei,Kamnitas}, primarily thanks to large-scale annotated training data. Unfortunately, obtaining such massive medical data is challenging; consequently, better training requires intensive Data Augmentation (DA) techniques, such as geometric/intensity transformations of original images~\cite{Ronneberger,Milletari}.
However, those transformed images intrinsically have a similar distribution with respect to the original ones, leading to limited performance improvement; thus, generating realistic (i.e., similar to the real image distribution) but completely new samples is essential to fill the real image distribution uncovered by the original dataset. In this context, Generative Adversarial Network (GAN)-based DA is promising, as it has shown excellent performance in computer vision, revealing good generalization ability.
Especially, SimGAN outperformed the state-of-the-art with $21\%$ improvement in eye-gaze estimation~\cite{Shrivastava}.

Also in medical imaging, realistic retinal image and Computed Tomography (CT) image generation have been tackled using adversarial learning ~\cite{Costa,Chuquicusma}; a very recent study reported performance improvement with synthetic training data in CNN-based liver lesion classification, using a small number of $64 \times 64$ CT images for GAN training~\cite{Frid-Adar}. However, GAN-based image generation using MRI, the most effective modality for soft-tissue acquisition, has not yet been reported due to the difficulties from low-contrast MR images, strong anatomical consistency, and intra-sequence variability; in our previous work~\cite{HAN}, we generated $64 \times 64$/$128 \times 128$ MR images using conventional GANs and even an expert physician failed to accurately distinguish between the real/synthetic images.

\begin{figure}[t]
  \centering
  \centerline{\includegraphics[width=12cm]{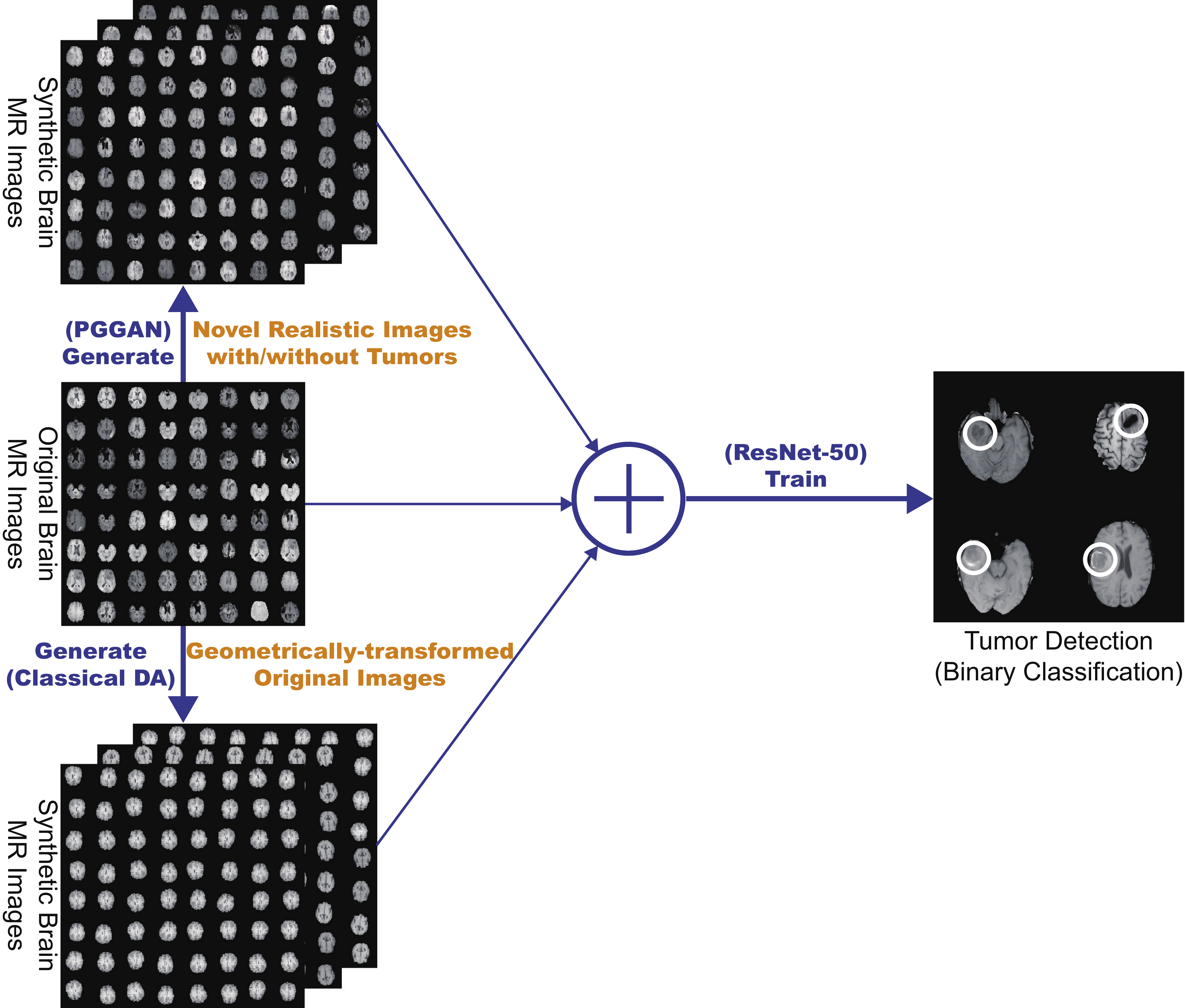}}
\caption{PGGAN-based DA for better tumor detection: the PGGANs method generates a number of realistic brain tumor/non-tumor MR images and the binary classifier uses them as additional training data.}
\label{fig1}
\vspace{-0.2in}
\end{figure}

So, how can we generate highly-realistic and original-sized $256\times256$ images, while maintaining clear tumor/non-tumor features using GANs?
Our aim is to generate GAN-based synthetic contrast-enhanced T1-weighted (T1c) brain MR images---the most commonly used sequence in tumor detection thanks to its high-contrast~\cite{Militello,rundoCMPB2017}---for CNN-based tumor detection. This computer-assisted brain tumor MRI analysis task is clinically valuable for better diagnosis, prognosis, and treatment~\cite{Havaei,Kamnitas}.
Generating $256\times256$ images is extremely challenging: (\textit{i}) GAN training is unstable with high-resolution inputs and severe artifacts appear due to strong consistency in brain anatomy; (\textit{ii}) brain tumors vary in size, location, shape, and contrast.
However, it is beneficial, because most CNN architectures adopt around $256\times256$ input sizes (e.g., Inception-ResNet-V2~\cite{Szegedy}: $299\times299$, ResNet-50~\cite{He}: $224\times224$) and we can achieve better results with original-sized image augmentation---towards this, we use Progressive Growing of GANs (PGGANs), a multi-stage generative training method~\cite{Karras}. Moreover, an expert physician evaluates the generated images' realism and tumor/non-tumor features \textit{via} the Visual Turing Test~\cite{Salimans}.
Using the synthetic images, our novel PGGAN-based DA approach achieves better performance in CNN-based tumor detection, when combined with classical DA (Fig.~\ref{fig1}).

\paragraph{Contributions.}
Our main contributions are as follows: 

\begin{itemize}
\item \textbf{MR Image Generation:} This research explains how to exploit MRI data to generate realistic and original-sized $256\times256$ whole brain MR images using PGGANs, while maintaining clear tumor/non-tumor features.
\item \textbf{MR Image Augmentation:} This study shows encouraging results on PGGAN-based DA, when combined with classical DA, for better tumor detection and other medical imaging tasks.
\end{itemize}

The rest of the manuscript is organized as follows: Sect. \ref{sec:back} introduces background on GANs; Sect. \ref{sec:method} describes our MRI dataset and PGGAN-based DA approach for tumor detection with its validations; experimental results are shown and analyzed in Sect. \ref{sec:results}; Sect. \ref{sec:conclusion} presents conclusion and future work.

\section{Generative Adversarial Networks}
\label{sec:back}
Originally proposed by Goodfellow \textit{et al.} in 2014~\cite{Goodfellow}, GANs have shown remarkable results in image generation~\cite{Zhu} relying on a two-player minimax game: a generator network aims at generating realistic images to fool a discriminator network that aims at distinguishing between the real/synthetic images. However, the two-player objective function leads to difficult training accompanying artificiality and mode collapse~\cite{Gulrajani}, especially with high resolution.
Deep Convolutional GAN (DCGAN)~\cite{Radford}, the most standard GAN, results in stable training on $64\times64$ images. In this context, several multi-stage generative training methods have been proposed: Composite GAN exploits multiple generators to separately generate different parts of an image~\cite{Kwak}; the PGGANs method adopts multiple training procedures from low resolution to high to incrementally generate a realistic image \cite{Karras}.

Recently, researchers applied GANs to medical imaging, mainly for image-to-image translation, such as segmentation~\cite{Xue}, super-resolution~\cite{Mahapatra}, and cross-modality translation~\cite{Nie}. Since GANs allow for adding conditional dependency on the input information (e.g., category, image, and text), they used such conditional GANs to produce the desired corresponding images. However, GAN-based research on generating large-scale synthetic training images is limited, while the biggest challenge in this field is handling small datasets.

Differently from a very recent DA work for $64 \times 64$ CT liver lesion Region of Interest (ROI) classification~\cite{Frid-Adar}, to the best of our knowledge, this is the first GAN-based whole MR image augmentation approach. This work also firstly uses PGGANs to generate $256 \times 256$ medical images. Along with classical transformations of real images, a completely different approach---generating novel realistic images using PGGANs---may become a clinical breakthrough.

\section{Materials and Methods}
\label{sec:method}
\subsection{BRATS 2016 Training Dataset}
This paper exploits a dataset of $240 \times 240$ T1c brain axial MR images containing $220$ High-Grade Glioma cases to train PGGANs with sufficient data and image resolution.
These MR images are extracted from the Multimodal Brain Tumor Image Segmentation Benchmark (BRATS) 2016 \cite{Menze}.

\subsection{PGGAN-based Image Generation}
\subsubsection{Data Preparation}
We select the slices from $\#30$ to $\#130$ among the whole $155$ slices to omit initial/final slices, since they convey a negligible amount of useful information and negatively affect the training of both PGGANs and ResNet-50. For tumor detection, our whole dataset ($220$ patients) is divided into: (\textit{i}) a training set ($154$ patients); (\textit{ii}) a validation set ($44$ patients); (\textit{iii}) a test set ($22$ patients). Only the training set is used for the PGGAN training to be fair. Since tumor/non-tumor annotations are based on 3D volumes, these labels are often incorrect/ambiguous on 2D slices; so, we discard (\textit{i}) tumor images tagged as non-tumor, (\textit{ii}) non-tumor images tagged as tumor, (\textit{iii}) unclear boundary images, and (\textit{iv}) too small/big images; after all, our datasets consist of:

\begin{itemize}
\item Training set ($5,036$ tumor/$3,853$ non-tumor images);
\item Validation set ($793$ tumor/$640$ non-tumor images);
\item Test set ($1,575$ tumor/$1,082$ non-tumor images).
\end{itemize}

\begin{figure}[t]
  \centering
  \centerline{\includegraphics[width=12cm]{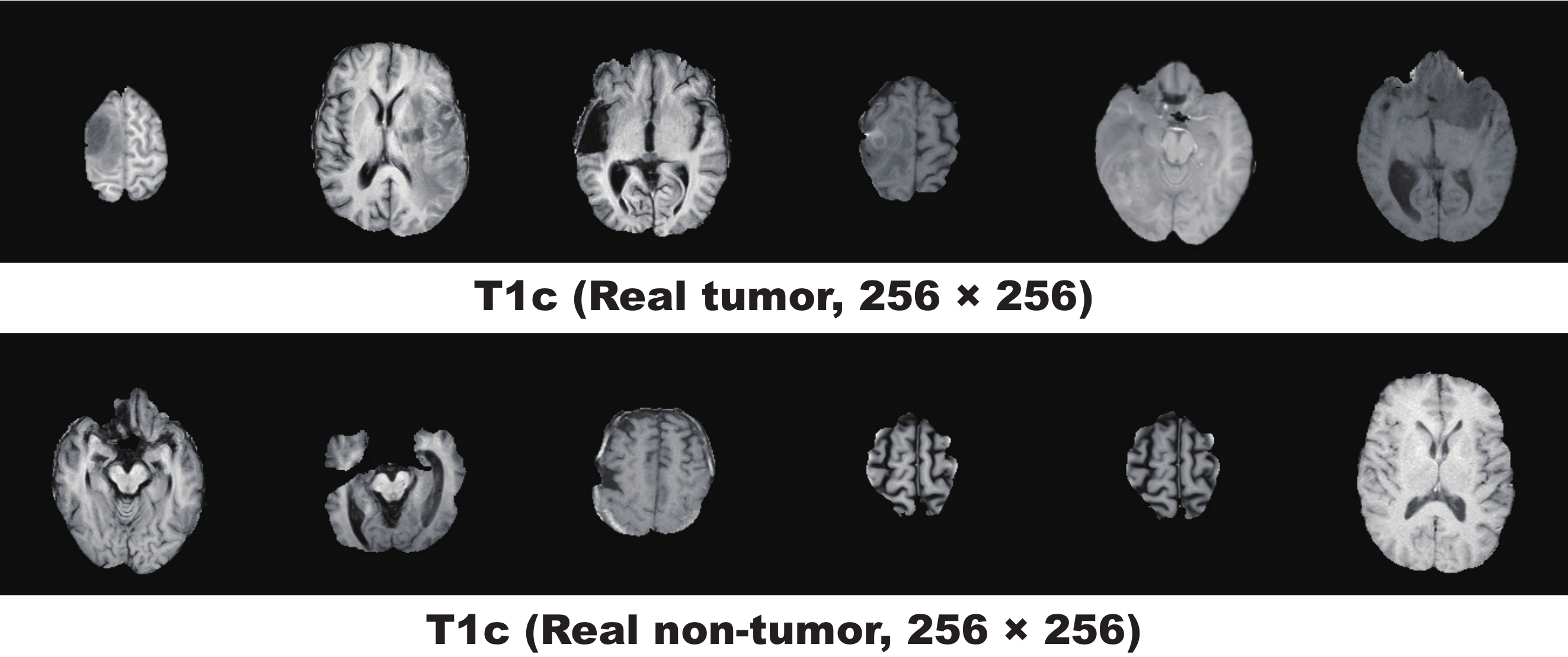}}
\caption{Example real $256 \times 256$ MR images used for PGGAN training.}
\label{fig2}
\end{figure}

\begin{figure}[H]
  \centering
  \centerline{\includegraphics[width=12cm]{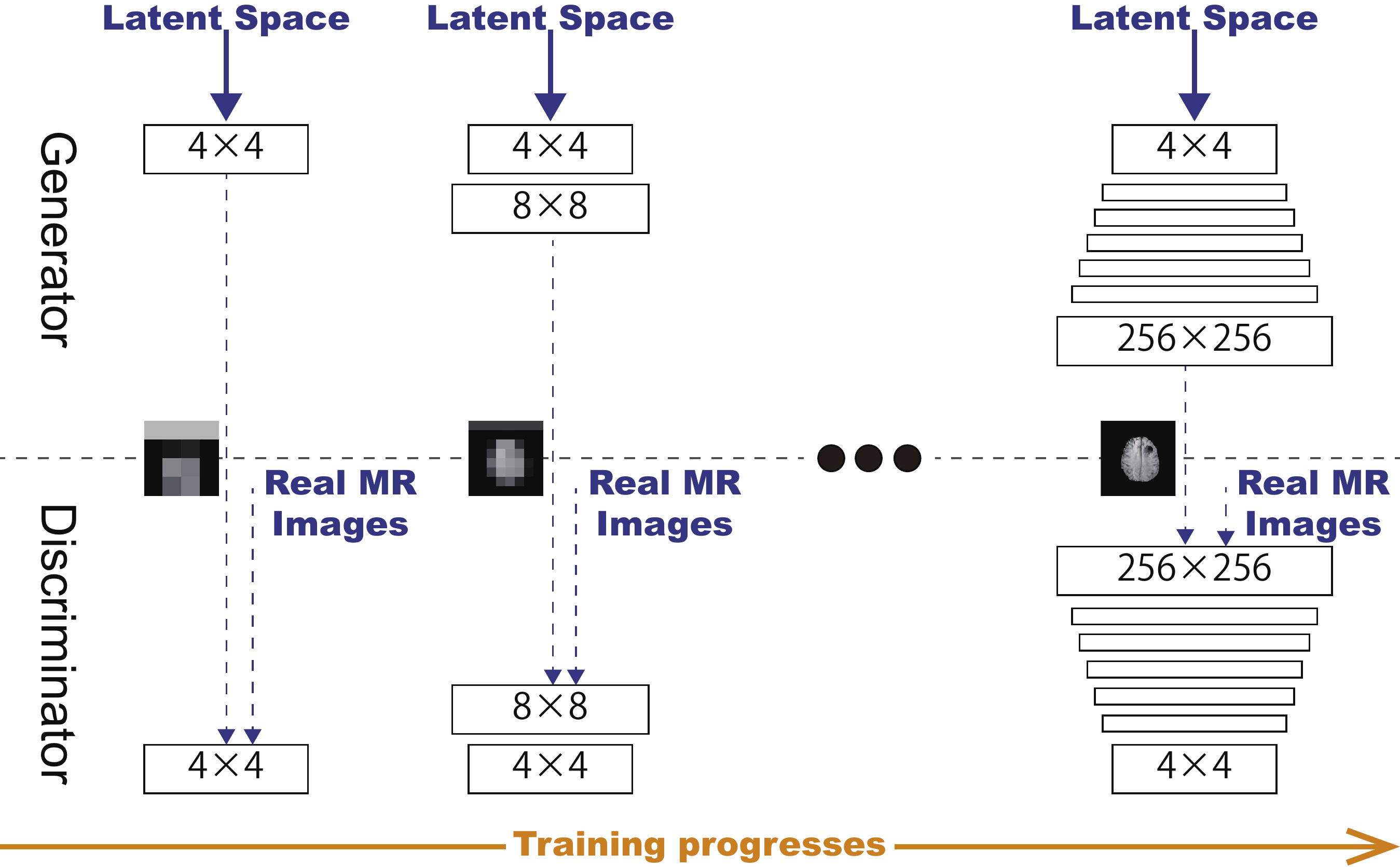}}
\caption{PGGANs architecture for synthetic $256 \times 256$ MR image generation.}
\label{fig3}
\end{figure}

The images from the training set are zero-padded to reach a power of $2$, $256 \times 256$ from $240 \times 240$ pixels for better PGGAN training. Fig.~\ref{fig2} shows examples of real MR images.

\subsubsection{PGGANs} is a novel training method for GANs with progressively growing generator and discriminator~\cite{Karras}: starting from low resolution, newly added layers model fine-grained details as training progresses. As Fig.~\ref{fig3} shows, we adopt PGGANs to generate highly-realistic and original-sized $256 \times 256$ brain MR images; tumor/non-tumor images are separately trained and generated.

\subsubsection{PGGAN Implementation Details} We use the PGGAN architecture with the Wasserstein loss using gradient penalty~\cite{Gulrajani}. Training lasts for $100$ epochs with a batch size of 16 and $1.0 \times 10^{-3}$ learning rate for Adam optimizer.

\subsection{Tumor Detection Using ResNet-50}
\subsubsection{Pre-processing} To fit ResNet-50's input size, we center-crop the whole images from $240 \times 240$ to $224 \times 224$ pixels.
\subsubsection{ResNet-50} is a residual learning-based CNN with $50$ layers~\cite{He}: unlike conventional learning unreferenced functions, it reformulates the layers as learning residual functions for sustainable and easy training. We adopt ResNet-50 to detect tumors in brain MR images, i.e., the binary classification of images with/without tumors.

\begin{figure}[t]
  \centering
  \centerline{\includegraphics[width=10cm]{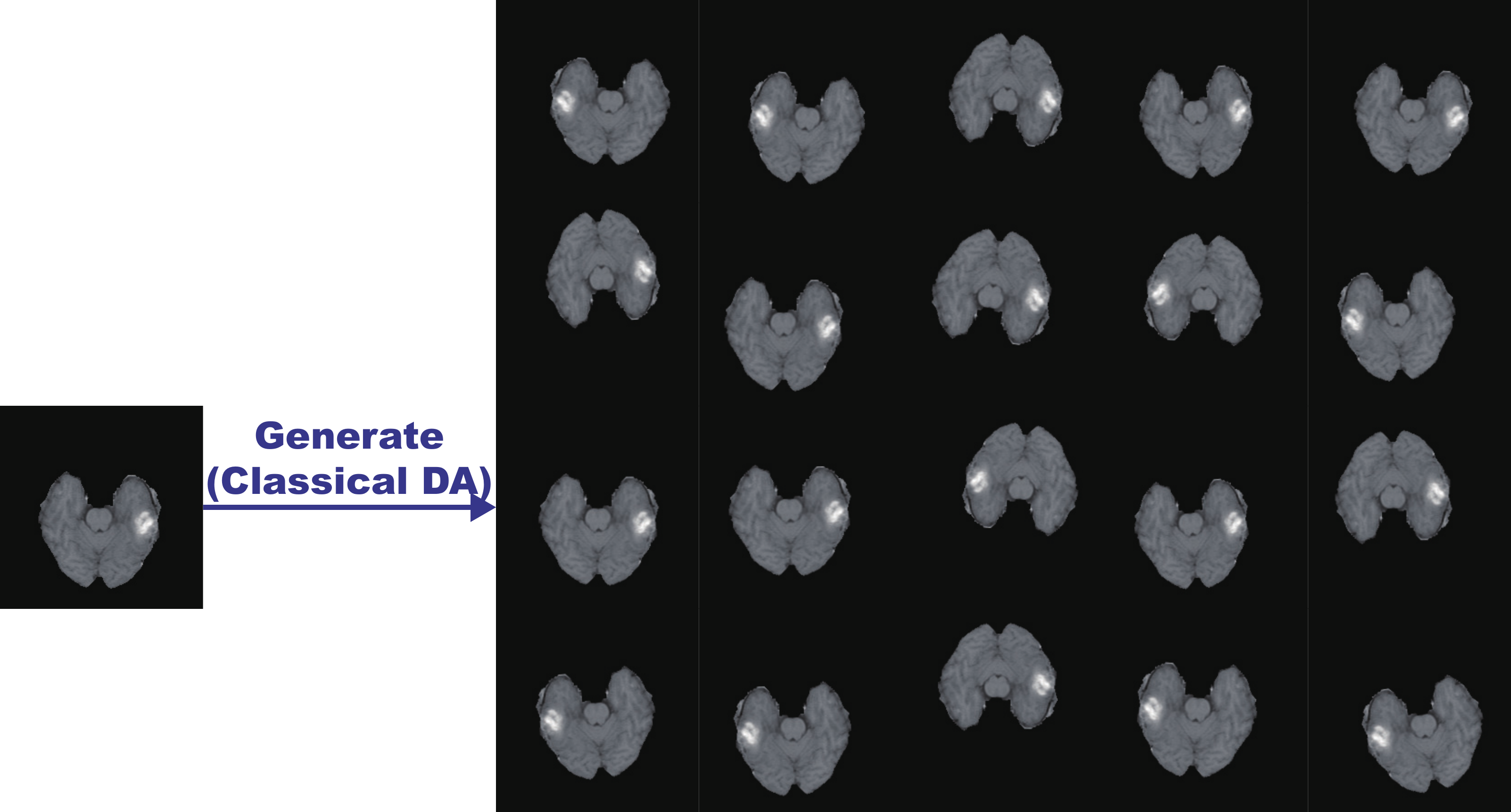}}
\caption{Example real MR image and its geometrically-transformed synthetic images.}
\label{fig4}
\end{figure}

To confirm the effect of PGGAN-based DA, the following classification results are compared: (\textit{i}) without DA, (\textit{ii}) with $200,000$ classical DA ($100,000$ for each class), (\textit{iii}) with $200,000$ PGGAN-based DA, and (\textit{iv}) with both $200,000$ classical DA and $200,000$ PGGAN-based DA; the classical DA adopts a random combination of horizontal/vertical flipping, rotation up to $10$ degrees, width/height shift up to $8\%$, shearing up to $8\%$, zooming up to $8\%$, and constant filling of points outside the input boundaries (Fig.~\ref{fig4}). For better DA, highly-unrealistic PGGAN-generated images are manually discarded.

\subsubsection{ResNet-50 Implementation Details}
We use the ResNet-50 architecture pre-trained on ImageNet with a dropout of $0.5$ before the final softmax layer, along with a batch size of $192$, $1.0 \times 10^{-3}$ learning rate for Adam optimizer, and early stopping of $10$ epochs.

\subsection{Clinical Validation Using the Visual Turing Test}
To quantitatively evaluate (\textit{i}) how realistic the PGGAN-based synthetic images are, (\textit{ii}) how obvious the synthetic images' tumor/non-tumor features are, we supply, in a random order, to an expert physician a random selection of:
\begin{itemize}
\item $50$ real tumor images;
\item $50$ real non-tumor images;
\item $50$ synthetic tumor images;
\item $50$ synthetic non-tumor images.
\end{itemize}

Then, the physician is asked to constantly classify them as both (\textit{i}) real/synthetic and (\textit{ii}) tumor/non-tumor, without previous training stages revealing which is real/synthetic and tumor/non-tumor; here, we only show successful cases of synthetic images, as we can discard failed cases for better data augmentation.
The so-called Visual Turing Test~\cite{Salimans} is used to probe the human ability to identify attributes and relationships in images, also in evaluating the visual quality of GAN-generated images~\cite{Shrivastava}.
Similarly, this applies to medical images in clinical environments~\cite{Chuquicusma,Frid-Adar}, wherein physicians' expertise is critical.

\subsection{Visualization Using t-SNE}
To visually analyze the distribution of both (\textit{i}) real/synthetic and (\textit{ii}) tumor/non-tumor images, we use t-Distributed Stochastic Neighbor Embedding (t-SNE)~\cite{Maaten} on a random selection of:
\begin{itemize}
\item $300$ real non-tumor images;
\item $300$ geometrically-transformed non-tumor images;
\item $300$ PGGAN-generated non-tumor images;
\item $300$ real tumor images;
\item $300$ geometrically-transformed tumor images;
\item $300$ PGGAN-generated tumor images.
\end{itemize}

Only 300 images per each category are selected for better visualization. t-SNE is a machine learning algorithm for dimensionality reduction to represent high-dimensional data into a lower-dimensional (2D/3D) space. It non-linearly adapts to input data using perplexity, which balances between the data's local and global aspects.

\subsubsection{t-SNE Implementation Details}
We use t-SNE with a perplexity of $100$ for 1,000 iterations to obtain a 2D visual representation.

\section{Results}
\label{sec:results}
This section shows how PGGANs generates synthetic brain MR images. The results include instances of synthetic images, their quantitative evaluation by an expert physician, and their influence on tumor detection.

\subsection{MR Images Generated by PGGANs}

\begin{figure}[t]
  \centering
  \centerline{\includegraphics[width=12cm]{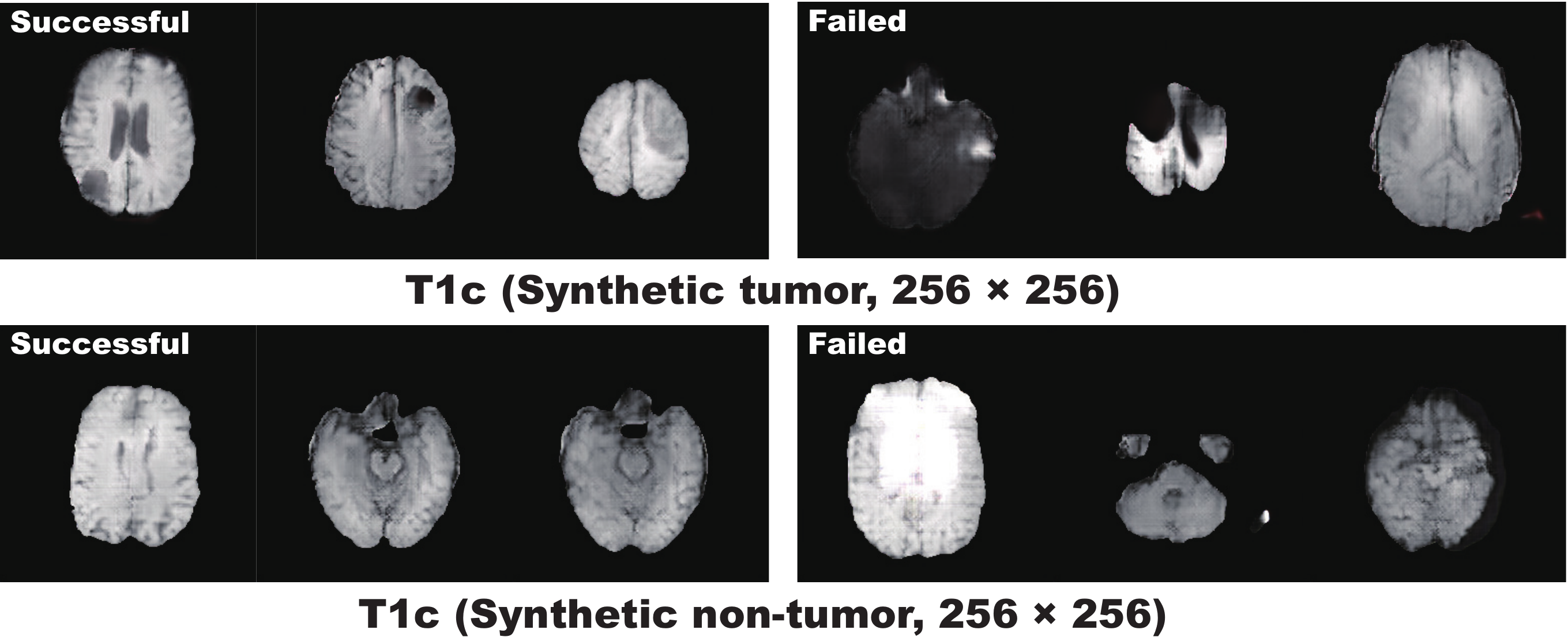}}
\caption{Example synthetic MR images yielded by PGGANs: (a) successful cases; (b) failed cases.}
\label{fig5}
\end{figure}

Fig.~\ref{fig5} illustrates examples of synthetic tumor/non-tumor images by PGGANs. In our visual confirmation, for about $75\%$ of cases, PGGANs successfully captures the T1c-specific texture and tumor appearance while maintaining the realism of the original brain MR images; however, for about $25\%$ of cases, the generated images lack clear tumor/non-tumor features or contain unrealistic features, such as hyper-intensity, gray contours, and odd artifacts. 

\begin{table*}[!t]
\caption{Binary classification results for detecting brain tumors with/without DA.}
\label{tab1}
\centering
\begin{scriptsize}
\begin{tabular}{lccccc}
\Hline\noalign{\smallskip}
\bfseries Experimental condition \ & \multicolumn{1}{c}{ Accuracy}  \ & Sensitivity \ & Specificity \\\noalign{\smallskip}\hline\noalign{\smallskip}
ResNet-50 (w/o DA) \ & $90.06\%$ \ & $85.27\%$ \ & $97.04\%$ \\
ResNet-50 (w/ 200k classical DA) \ & $90.70\%$ \ & $88.70\%$ \ & $93.62\%$ \\
ResNet-50 (w/ 200k PGGAN-based DA) \ & $62.02\%$ \ & $\mathbf{99.94}\%$ \ & $6.84\%$ \\
ResNet-50 (w/ 200k classical DA + 200k PGGAN-based DA) \ & $\mathbf{91.08}\%$ \ & $86.60\%$ \ & $\mathbf{97.60}\%$ \\
\noalign{\smallskip}\Hline\noalign{\smallskip}
\end{tabular}
\end{scriptsize}
\end{table*}

\subsection{Tumor Detection Results}

Table~\ref{tab1} shows the classification results for detecting brain tumors with/without DA techniques. As expected, the test accuracy improves by $0.64\%$ with the additional $200,000$ geometrically-transformed images for training. When only the PGGAN-based DA is applied, the test accuracy decreases drastically with almost 100\% of sensitivity and $6.84\%$ of specificity, because the classifier recognizes the synthetic images' prevailed unrealistic features as tumors, similarly to anomaly detection.

However, surprisingly, when it is combined with the classical DA, the accuracy increases by $1.02\%$ with higher sensitivity and specificity; this could occur because the PGGAN-based DA fills the real image distribution uncovered by the original dataset, while the classical DA provides the robustness on training for most cases.

\begin{table*}[!t]
\caption{Visual Turing Test results by a physician for classifying Real (\textit{R}) vs Synthetic (\textit{S}) images and Tumor (\textit{T}) vs Non-tumor (\textit{N}) images.} 
\label{tab2}
\centering
\begin{footnotesize}
\begin{tabular}{ccccc}
\Hline\noalign{\smallskip}
\multicolumn{1}{c}{Real/Synthetic Classification}  \ \ \ &  \textit{R} as \textit{R} \ \ \ &  \textit{R} as \textit{S} \ \ \ &  \textit{S} as \textit{R} \ \ \ &  \textit{S} as \textit{S} \\\noalign{\smallskip}\hline\noalign{\smallskip}
$78.5\%$ \ \ \ & $58$ \ \ \ & $42$ \ \ \ & $1$ \ \ \ & $99$ \\
\noalign{\smallskip}\hline\hline\noalign{\smallskip}
\multicolumn{1}{c}{Tumor/Non-tumor Classification} \ \ \ &  \textit{T} as \textit{T} \ \ \ &  \textit{T} as \textit{N} \ \ \ &  \textit{N} as \textit{T} \ \ \ &  \textit{N} as \textit{N}\\
\noalign{\smallskip}\hline\noalign{\smallskip}
$90.5\%$ \ \ \ & $82$ \ \ \ & $18$ (\textit{R}: $5$, \textit{S}: $13$) \ \ \ & $1$ (\textit{S}: $1$) \ \ \ & $99$\\
\noalign{\smallskip}\Hline\noalign{\smallskip}
\end{tabular}
\end{footnotesize}
\end{table*}

\subsection{Visual Turing Test Results}
Table~\ref{tab2} shows the confusion matrix for the Visual Turing Test. Differently from our previous work on GAN-based $64 \times 64$/$128 \times 128$ MR image generation, the expert physician easily recognizes $256 \times 256$ synthetic images \cite{HAN}, while tending also to classify real images as synthetic; this can be attributed to high resolution associated with more difficult training and detailed appearance, making artifacts stand out, which is coherent to the ResNet-50's low tumor detection accuracy with only the PGGAN-based DA. Generally, the physician's tumor/non-tumor classification accuracy is high and the synthetic images successfully capture tumor/non-tumor features. However, unlike non-tumor images, the expert recognizes a considerable number of tumor images as non-tumor, especially on the synthetic images; this results from the remaining real images' ambiguous annotation, which is amplified in the synthetic images trained on them.

\begin{figure}[t]
  \centering
  \centerline{\includegraphics[width=12cm]{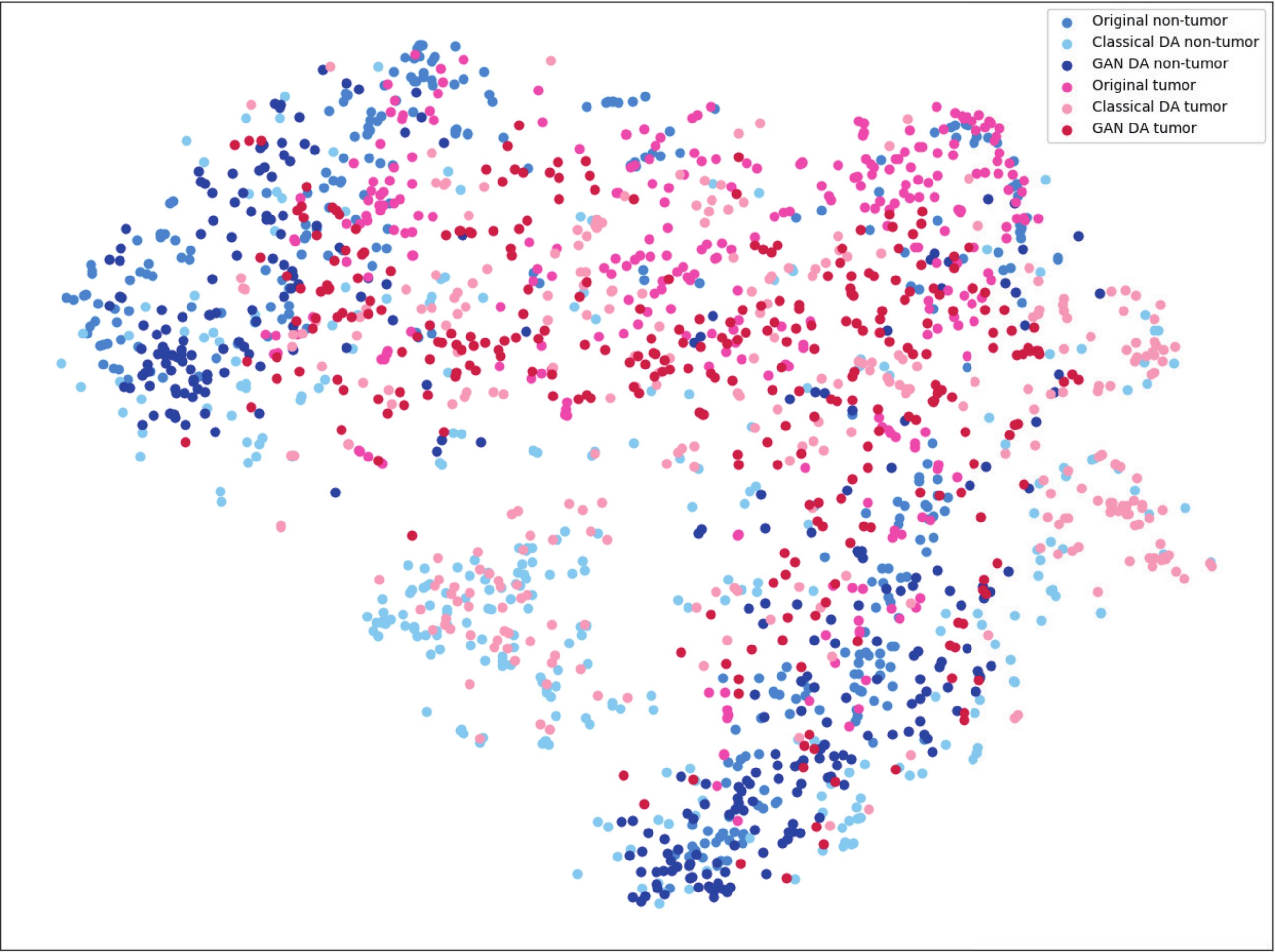}}
\caption{t-SNE result on six categories, with $300$ images per each category: (a) real tumor/non-tumor images; (b) geometrically-transformed tumor/non-tumor images; (c) PGGAN-generated tumor/non-tumor images.}
\label{fig7}
\end{figure}

\subsection{t-SNE Result}
As presented in Fig.~\ref{fig7}, tumor/non-tumor images' distribution shows a tendency that non-tumor images locate from top left to bottom right and tumor images locate from top right to center, while the distinction is unclear with partial overlaps. Classical DA covers a wide range, including zones without any real/GAN-generated images, but tumor/non-tumor images often overlap there.
Meanwhile, PGGAN-generated images concentrate differently from real images, while showing more frequent overlaps than the real ones; this probably derives from those synthetic images with unsatisfactory realism and tumor/non-tumor features.

\section{Conclusion}
\label{sec:conclusion}
Our preliminary results show that PGGANs can generate original-sized $256 \times 256$ realistic brain MR images and achieve higher performance in tumor detection, when combined with classical DA.
This occurs because PGGANs' multi-stage image generation obtains good generalization and synthesizes images with the real image distribution unfilled by the original dataset. However, considering the Visual Turing Test and t-SNE results, yet unsatisfactory realism with high resolution strongly limits DA performance, so we plan to (\textit{i}) generate only realistic images, and then (\textit{ii}) refine synthetic images more similar to the real image distribution.

For (\textit{i}), we can map an input random vector onto each training image~\cite{Schlegl} and generate images with suitable vectors, to control the divergence of generated images; virtual adversarial training could be also integrated to control the output distribution. Moreover, (\textit{ii}) can be achieved by GAN/VAE-based image-to-image translation, such as Unsupervised Image-to-Image Translation Networks~\cite{Liu}, considering SimGAN's remarkable performance improvement after refinement~\cite{Shrivastava}. Moreover, we should further avoid real images with ambiguous/inaccurate annotation for better tumor detection.

Overall, our novel PGGAN-based DA approach sheds light on diagnostic and prognostic medical applications, not limited to tumor detection; future studies are needed to extend our encouraging results.
%
%
%
%
\section*{Acknowledgment}
This work was partially supported by the Graduate Program for Social ICT Global Creative Leaders of The University of Tokyo by JSPS.

%
%

\bibliographystyle{splncs} 
\bibliography{biblio}

\begin{thebibliography}{10}

\bibitem{Rundo}
Rundo, L., Militello, C., Russo, G., Vitabile, S., Gilardi, M.C., Mauri, G.:
\newblock {GTVcut} for neuro-radiosurgery treatment planning: an {MRI} brain
  cancer seeded image segmentation method based on a cellular automata model.
\newblock Nat. Comput. \textbf{17}(3) (2018)  521--536

\bibitem{rundo2016WIRN}
Rundo, L., Militello, C., Vitabile, S., Russo, G., Pisciotta, P., Marletta, F.,
  Ippolito, M., D’Arrigo, C., Midiri, M., Gilardi, M.C.:
\newblock Semi-automatic brain lesion segmentation in {Gamma Knife} treatments
  using an unsupervised fuzzy c-means clustering technique.
\newblock In: Advances in Neural Networks: Computational Intelligence for ICT.
  Volume~54 of Smart Innovation, Systems and Technologies.
\newblock Springer (2016)  15--26

\bibitem{Bevilacqua}
Bevilacqua, V., Brunetti, A., Cascarano, G.D., Palmieri, F., Guerriero, A.,
  Moschetta, M.:
\newblock A deep learning approach for the automatic detection and segmentation
  in autosomal dominant polycystic kidney disease based on {Magnetic Resonance}
  images.
\newblock In: Proc. International Conference on Intelligent Computing (ICIP),
  Springer (2018)  643--649

\bibitem{Brunetti}
Brunetti, A., Carnimeo, L., Trotta, G.F., Bevilacqua, V.:
\newblock Computer-assisted frameworks for classification of liver, breast and
  blood neoplasias via neural networks: a survey based on medical images.
\newblock Neurocomputing (2018)

\bibitem{Havaei}
Havaei, M., Davy, A., Warde-Farley, D., Biard, A., Courville, A., Bengio, Y.,
  et~al.:
\newblock Brain tumor segmentation with deep neural networks.
\newblock Med. Image Anal. \textbf{35} (2017)  18--31

\bibitem{Kamnitas}
Kamnitsas, K., Ledig, C., Newcombe, V.F.J., Simpson, J.P., Kane, A.D., Menon,
  D.K.,  et~al.:
\newblock Efficient multi-scale {3D} {CNN} with fully connected {CRF} for
  accurate brain lesion segmentation.
\newblock Med. Image Anal. \textbf{36} (2017)  61--78

\bibitem{Ronneberger}
Ronneberger, O., Fischer, P., Brox, T.:
\newblock {U-Net}: Convolutional networks for biomedical image segmentation.
\newblock In: Proc. International Conference on Medical Image Computing and
  Computer-Assisted Intervention (MICCAI). (2015)  234--241

\bibitem{Milletari}
Milletari, F., Navab, N., Ahmadi, S.:
\newblock {V-Net}: Fully convolutional neural networks for volumetric medical
  image segmentation.
\newblock In: Proc. International Conference on 3D Vision (3DV), IEEE (2016)
  565--571

\bibitem{Shrivastava}
Shrivastava, A., Pfister, T., Tuzel, O., Susskind, J., Wang, W., Webb, R.:
\newblock Learning from simulated and unsupervised images through adversarial
  training.
\newblock In: Proc. Conference on Computer Vision and Pattern Recognition
  (CVPR), IEEE (2017)  2107--2116

\bibitem{Costa}
Costa, P., Galdran, A., Meyer, M.I., Niemeijer, M., Abr\`amoff, M., Mendonça,
  A.M., Campilho, A.:
\newblock End-to-end adversarial retinal image synthesis.
\newblock IEEE Trans. Med. Imaging \textbf{37}(3) (2018)  781--791

\bibitem{Chuquicusma}
Chuquicusma, M.J.M., Hussein, S., Burt, J., Bagci, U.:
\newblock How to fool radiologists with generative adversarial networks? a
  visual {Turing} test for lung cancer diagnosis.
\newblock In: Proc. International Symposium on Biomedical Imaging (ISBI), IEEE
  (2018)  240--244

\bibitem{Frid-Adar}
Frid-Adar, M., Diamant, I., Klang, E., Amitai, M., Goldberger, J., Greenspan,
  H.:
\newblock {GAN}-based synthetic medical image augmentation for increased {CNN}
  performance in liver lesion classification.
\newblock Neurocomputing \textbf{321} (2018)  321--331

\bibitem{HAN}
Han, C., Hayashi, H., Rundo, L., Araki, R., Shimoda, W., Muramatsu, S.,
  et~al.:
\newblock {GAN}-based synthetic brain {MR} image generation.
\newblock In: Proc. International Symposium on Biomedical Imaging (ISBI), IEEE
  (2018)  734--738

\bibitem{Militello}
Militello, C., Rundo, L., Vitabile, S.,  et~al.:
\newblock Gamma knife treatment planning: {MR} brain tumor segmentation and
  volume measurement based on unsupervised fuzzy c-means clustering.
\newblock Int. J. Imaging Syst. Technol. \textbf{25}(3) (2015)  213--225

\bibitem{rundoCMPB2017}
Rundo, L., Stefano, A., Militello, C., Russo, G., Sabini, M.G., D'Arrigo, C.,
  Marletta, F., Ippolito, M., Mauri, G., Vitabile, S., Gilardi, M.C.:
\newblock A fully automatic approach for multimodal {PET} and {MR} image
  segmentation in {Gamma Knife} treatment planning.
\newblock Comput. Methods Programs Biomed. \textbf{144} (2017)  77--96

\bibitem{Szegedy}
Szegedy, C., Ioffe, S., Vanhoucke, V., Alemi, A.A.:
\newblock Inception-v4, inception-resnet and the impact of residual connections
  on learning.
\newblock In: Proc. AAAI Conference on Artificial Intelligence (AAAI). (2017)

\bibitem{He}
He, K., Zhang, X., Ren, S., Sun, J.:
\newblock Deep residual learning for image recognition.
\newblock In: Proc. Conference on Computer Vision and Pattern Recognition
  (CVPR), IEEE (2016)  770--778

\bibitem{Karras}
Karras, T., Aila, T., Laine, S., Lehtinen, J.:
\newblock Progressive growing of {GAN}s for improved quality, stability, and
  variation.
\newblock In: Proc. International Conference on Learning Representations
  (ICLR), arXiv preprint arXiv:1710.10196v3. (2018)

\bibitem{Salimans}
Salimans, T., Goodfellow, I., Zaremba, W., Cheung, V., Radford, A., Chen, X.:
\newblock Improved techniques for training {GAN}s.
\newblock In: Advances in Neural Information Processing Systems (NIPS). (2016)
  2234--2242

\bibitem{Goodfellow}
Goodfellow, I., Pouget-Abadie, J., Mirza, M., Xu, B., Warde-Farley, D., Ozair,
  S.,  et~al.:
\newblock Generative adversarial nets.
\newblock In: Advances in Neural Information Processing Systems (NIPS). (2014)
  2672--2680

\bibitem{Zhu}
Zhu, J.Y., Park, T., Isola, P., Efros, A.A.:
\newblock Unpaired image-to-image translation using cycle-consistent
  adversarial networks.
\newblock In: Proc. International Conference on Computer Vision (ICCV), IEEE
  (2017)  2242--2251

\bibitem{Gulrajani}
Gulrajani, I., Ahmed, F., Arjovsky, M., Dumoulin, V., Courville, A.C.:
\newblock Improved training of {Wasserstein} {GANs}.
\newblock In: Advances in Neural Information Processing Systems. (2017)
  5769--5779

\bibitem{Radford}
Radford, A., Metz, L., Chintala, S.:
\newblock Unsupervised representation learning with deep convolutional
  generative adversarial networks.
\newblock In: Proc. International Conference on Learning Representations
  (ICLR), arXiv preprint arXiv:1511.06434. (2016)

\bibitem{Kwak}
Kwak, H., Zhang, B.:
\newblock Generating images part by part with composite generative adversarial
  networks.
\newblock arXiv preprint arXiv:1607.05387 (2016)

\bibitem{Xue}
Xue, Y., Xu, T., Zhang, H., Long, L.R., Huang, X.:
\newblock {SegAN}: Adversarial network with multi-scale ${L}_{1}$ loss for
  medical image segmentation.
\newblock Neuroinformatics \textbf{16}(3--4) (2018)  383–392

\bibitem{Mahapatra}
Mahapatra, D., Bozorgtabar, B., Hewavitharanage, S., Garnavi, R.:
\newblock Image super resolution using generative adversarial networks and
  local saliency maps for retinal image analysis.
\newblock In: Proc. International Conference on Medical Image Computing and
  Computer-Assisted Intervention (MICCAI). (2017)  382--390

\bibitem{Nie}
Nie, D., Trullo, R., Lian, J., Petitjean, C., Ruan, S., Wang, Q., Shen, D.:
\newblock Medical image synthesis with context-aware generative adversarial
  networks.
\newblock In: Proc. International Conference on Medical Image Computing and
  Computer-Assisted Intervention (MICCAI). (2017)  417--425

\bibitem{Menze}
Menze, B.H., Jakab, A., Bauer, S.,  et~al.:
\newblock The multimodal brain tumor image segmentation benchmark {(BRATS)}.
\newblock IEEE Trans. Med. Imaging \textbf{34}(10) (2015)  1993--2024

\bibitem{Maaten}
van~der Maaten, L., Hinton, G.:
\newblock Visualizing data using {t-SNE}.
\newblock J. Mach. Learn. Res. \textbf{9} (2008)  2579--2605

\bibitem{Schlegl}
Schlegl, T., Seeb{\"o}ck, P., Waldstein, S.M., Schmidt-Erfurth, U., Langs, G.:
\newblock Unsupervised anomaly detection with generative adversarial networks
  to guide marker discovery.
\newblock In: Proc. International Conference on Information Processing in
  Medical Imaging (IPMI). (2017)  146--157

\bibitem{Liu}
Liu, M.Y., Breuel, T., Kautz, J.:
\newblock Unsupervised image-to-image translation networks.
\newblock In: Advances in Neural Information Processing Systems (NIPS). (2017)
  700--708

\end{thebibliography}

\end{document}